\documentclass[10pt,twocolumn,letterpaper]{article}

\usepackage{cvpr}
\usepackage{times}
\usepackage{epsfig}
\usepackage{graphicx}
\usepackage{amsmath}
\usepackage{amssymb}
\usepackage{subcaption} 
\usepackage{ mathrsfs }
\usepackage{multirow}

\usepackage[pagebackref=true,breaklinks=true,letterpaper=true,colorlinks,bookmarks=false]{hyperref}

\cvprfinalcopy % *** Uncomment this line for the final submission

\def\cvprPaperID{} % *** Enter the CVPR Paper ID here

\ifcvprfinal\fi
\begin{document}
%%%%%%%%% TITLE
\title{Toward Streaming Synapse Detection with Compositional ConvNets}

\author{Shibani Santurkar \ David Budden \ Alexander Matveev \ Heather Berlin \ \\ Hayk Saribekyan \ Yaron Meirovitch \ Nir Shavit\\
Massachusetts Institute of Technology\\
{\tt\small $\left\{\text{shibani,budden,amatveev,hberlin,hayks,yaronm,shanir}\right\}$@mit.edu}}
%\and
%David Budden\\
%Massachusetts Institute of Technology\\
%{\tt\small budden@csail.mit.edu}
%\and
%Alexander Matveev\\
%Massachusetts Institute of Technology\\
%{\tt\small amatveev@csail.mit.edu}
%\and
%Heather Berlin\\
%Massachusetts Institute of Technology\\
%{\tt\small hberlin@mit.edu}
%\and
%Hayk Saribekyan\\
%Massachusetts Institute of Technology\\
%{\tt\small hayks@mit.edu}
%\and
%Yaron Meirovitch\\
%Massachusetts Institute of Technology\\
%{\tt\small yaronm@mit.edu}
%\and
%Nir Shavit\\
%Massachusetts Institute of Technology\\
%{\tt\small shanir@csail.mit.edu}
%}

\maketitle
%\thispagestyle{empty}

%%%%%%%%% ABSTRACT
\begin{abstract}
Connectomics is an emerging field in neuroscience that aims to reconstruct the $3-$dimensional morphology of neurons from electron microscopy (EM) images. Recent studies have successfully demonstrated the use of convolutional neural networks (ConvNets) for segmenting cell membranes to individuate neurons. However, there has been comparatively little success in high-throughput identification of the intercellular synaptic connections required for deriving connectivity graphs. 

In this study, we take a compositional approach to segmenting synapses, modeling them explicitly as an intercellular cleft co-located with an asymmetric vesicle density along a cell membrane. Instead of requiring a deep network to learn all natural combinations of this compositionality, we train lighter networks to model the simpler marginal distributions of membranes, clefts and vesicles from just $100$ electron microscopy samples. These feature maps are then combined with simple rules-based heuristics derived from prior biological knowledge.

Our approach to synapse detection is both more accurate than previous state-of-the-art ($7$\% higher recall and $5$\% higher F1-score) and yields a $20$-fold speed-up compared to the previous fastest implementations. We demonstrate by reconstructing the first complete, directed connectome from the largest available anisotropic microscopy dataset ($245$ GB) of mouse somatosensory cortex (S1) in just $9.7\,$hours on a single shared-memory CPU system. We believe that this work marks an important step toward the goal of a microscope-pace streaming connectomics pipeline.
\end{abstract}
\section{Introduction}
Rigorous studies of neural circuits in mammals could uncover motifs underlying information processing and neuropathies at the core of disease \cite{lichtman2011big, pecca2011shank3, baloyannis2012vascular,zikopoulos2010changes}. Traditionally these studies involved the painstaking manual tracing of individual neurons through microscopy data \cite{kasthuri2009brain,lee2016anatomy,morgan2016fuzzy}. In the modern connectomics field, this manual effort has been largely replaced by high-throughput pipelines responsible for automated segmentation and morphological reconstruction of neurons from nanometer-scale electron microscopy (EM) \cite{lee2015recursive, ronneberger2015u, ciresan2012deep, seymour, matveev2016}. Automation is critical for any large-scale investigation of neuron organization, as even a humble $1\,mm^3$ volume of tissue contains many petabytes of EM data at the necessary resolution.

Of course, reconstructing 3-dimensional models of neuron ``skeletons" is only a partial solution to the larger goal of constructing neuron connectivity maps. Individual neurons are very densely packed but their connectivity is comparatively sparse, thus deriving connectivity is more complex than simply identifying the cell membranes of adjacent neurons. In recent years, several frameworks have been published that segment neurons with near-human accuracy \cite{lee2015recursive, ronneberger2015u, ciresan2012deep, seymour, matveev2016}. However, there has been substantially less success in identifying synapses -- the junctions whereby neurons connect and communicate. In the context of a neural connectivity graph, this is equivalent to identifying nodes but ignoring the edges between them.

Synapses are inherently compositional, identifiable in EM by a darkened cleft between adjacent neurons flanked by an asymmetric density of vesicles (neurotransmitter-carrying organelles). Accordingly, most previous approaches to synapse detection have used classifiers (typically random forests) with hand-crafted features to leverage this prior knowledge. Early examples include the work of Kreshuk \emph{et al.}, which captured voxel geometry and texture \cite{kreshuk2011automated}; and Becker \emph{et al.}, who extended this work to capture 3-dimensional contextual cues \cite{becker2013learning}. Although these approaches worked well on the datasets for which they were trained, they failed to generalize beyond the specific contrast and anisotropy ($4\times4\times45$ nm) of that dataset~\cite{bock2011network}.
\label{sec: introduction}
\begin{figure}[!t]
	\begin{center}
		\includegraphics[width=3.25in]{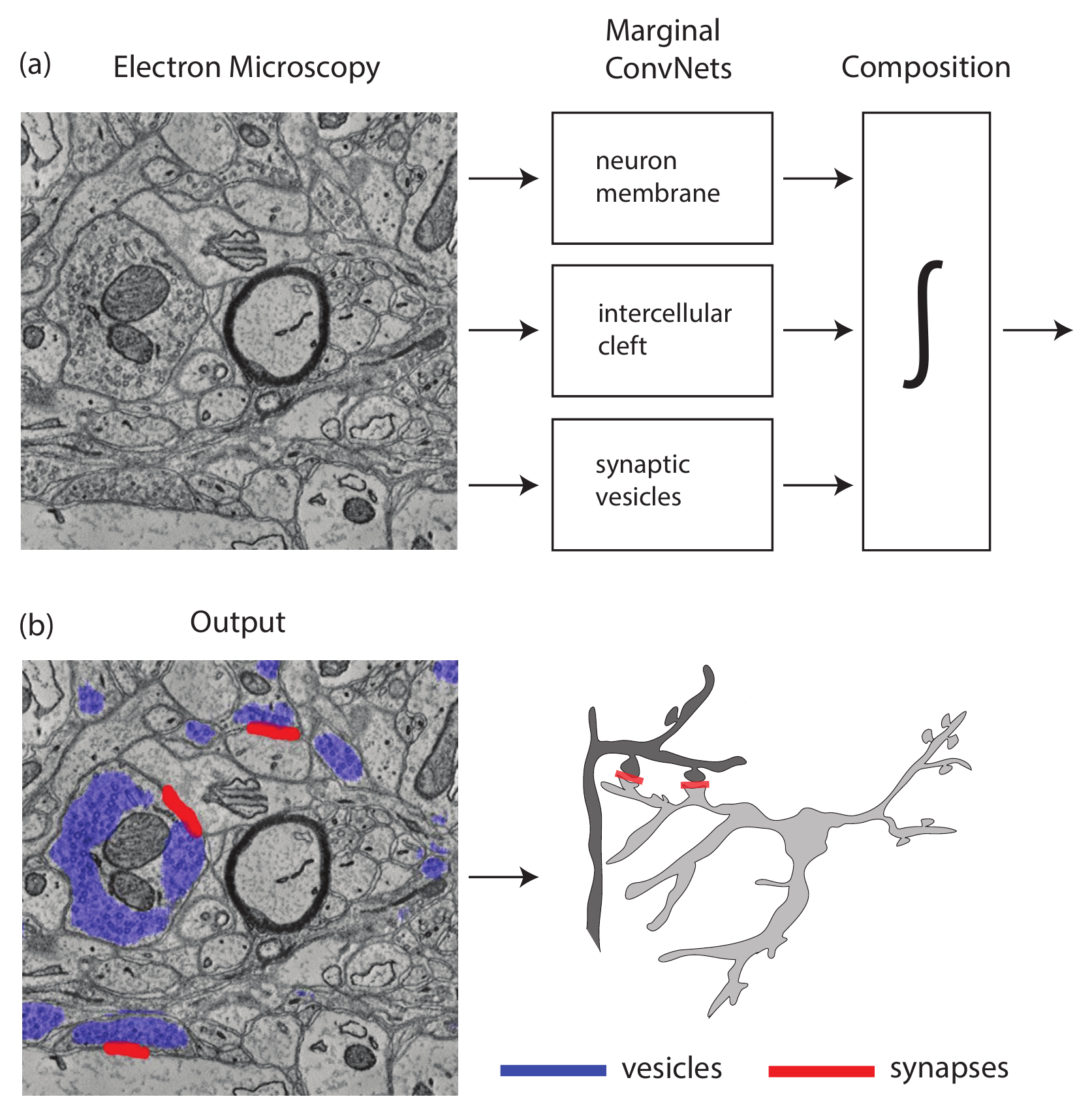}
	\end{center}
	\caption{(a) Workflow summary of our synapse detection. Raw electron microscopy (EM) data is streamed into lightweight parallel ConvNets, each trained to recognize a specific feature\textemdash one of neuron membranes, intercellular clefts or synaptic vesicles. (b) The output probability maps of each of these marginal features are then composed using prior biological knowledge to identify synapses (red). The asymmetric density of vesicles (blue) can also be leveraged to infer directionality of connectivity graphs.} %Vesicle clusters without adjoining synapses in the output EM correspond to synapses in neighboring slices.}
	\label{fig:workflow}
\end{figure}

In recent years, deep convolutional networks (ConvNets) have become a de facto standard for image classification \cite{krizhevsky2012imagenet} and semantic segmentation \cite{long2015fully}. When provided with sufficient training data, ConvNets typically outperform random forest and other classical machine learning approaches that are dependent on hand-crafted features. It is perhaps then unsurprising that the most accurate framework for synapse detection in previous literature is a ConvNet (Vesicle-CNN) by Roncal \emph{et al.}, which is trained to segment synapses directly from EM data without any prior knowledge of clefts, vesicles and their compositionality \cite{roncal2015vesicle}. The authors also published a much faster yet less accurate random forest classifier (Vesicle-RF), which outperforms previous approaches by explicitly modeling properties of synaptic connections.
%a post-synaptic density property of inter-cellular clefts.

The major issue with ConvNets (including Vesicle-CNN) is that they are prohibitively slow. Biologically meaningful volumes of neural tissue take months-to-years to image, thus a critical step toward the goals of connectomics is developing a computational pipeline that can process streaming data at microscope pace \cite{matveev2016,lichtman2014big}. The authors of Vesicle-CNN reported a processing time of $400\,$hours per GB of EM \cite{roncal2015vesicle}, i.e., several orders-of-magnitude slower than the TB/hr pace of modern multi-beam electron microscopes. Their lightweight Vesicle-RF was both substantially faster yet less accurate, making it impractical for connectomics where reliable synapse identification is needed for detecting enrichment of neural motifs.

In this study we present a new synapse detection framework that attempts to distill the advantages of both accurate ConvNets and faster knowledge-driven techniques. Rather than training a deep ConvNet to segment synapses directly from EM samples, we train much lighter networks to learn the marginal distributions of neuron membranes, intercellular clefts and synaptic vesicles. These features are then explicitly composed with simple rules that follow directly from prior biological knowledge.
The resulting system (see Figure~\ref{fig:workflow}) is both faster and more accurate compared to state-of-the art\textemdash $11000\times$ speed-up and $+3$\% F1-score over the accurate Vesicle-CNN and $20\times$ speed-up and $+5$\% F1-score over the faster yet less accurate Vesicle-RF. 
%The resulting system is both more accurate ($+3$\% F1-score) than Vesicle-CNN and $20 \times$ faster than the substantially less accurate Vesicle-RF. 

We apply our system to reconstruct the first complete, directed connectome from the largest available anisotropic EM dataset of mouse somatosensory cortex (S1 \cite{kasthuri2009brain}). This $245\,$GB reconstruction took just $9.7\,$hours on one multicore CPU, end-to-end from EM to connectivity graphs. Compare this to the previous largest $56\,$GB reconstruction, which needed $3\,$weeks to segment neuron membranes (on a farm of $27$ Titan X GPUs) and a further $39\,$hours to identify synapses using Vesicle-RF on a $100-$core CPU cluster \cite{roncal2015automated}.
\section{Compositional ConvNets}
\label{sec: algorithm}

In this section we describe our synapse detection system, which is both faster and more accurate than state-of-the-art and able to be run on a single multicore CPU server. This system is comprised of two main parts: (1) a bank of three lightweight ConvNets that can be deployed in parallel to segment membranes, clefts and synaptic vesicles; and (2) a rules-based module that integrates these three feature maps to identify synapses. The intuition behind this approach is to constrain the space in which our ConvNets are optimized by leveraging prior biological knowledge on synapse compositionality, i.e. the network simply has less to learn. This both avoids the computational burden associated with the traditional ``deeper is better" ConvNet paradigm while empirically improving synapse detection accuracy across the $245\,$GB S1 dataset. 

\subsection{Marginal Segmentation}
\label{sec:marg}
Deep ConvNets of small kernels have become a de facto standard for image classification tasks, largely motivated by the success of AlexNet \cite{krizhevsky2012imagenet} and progressively deeper networks \cite{simonyan2014very, szegedy2015going, he2015deep} in the annual ImageNet classification challenge \cite{ILSVRC15}. More recently ConvNets have been used to solve semantic segmentation tasks \cite{long2015fully}, which require assigning a class label or probability to each pixel of the output feature map. There have been several successful examples of ConvNets being applied successfully to segmentation of neuron membranes from EM data \cite{lee2015recursive, ronneberger2015u, ciresan2012deep, seymour,meirovitch2016multi}. In the context of reconstructing connectivity graphs, this is akin to defining the set of nodes but not edges.

In our system, we deploy lightweight ConvNets to learn the marginal distributions of (a) neuron membranes, (b) intercellular clefts and (c) synaptic vesicles. Our approach to segmenting neuron membranes is essentially equivalent to prior work in \cite{seymour}. The network is presented with $1024\times1024$ EM images at $6\,$nm resolution and associated ground-truth segmentations, and learns to produce an output segmentation in the form of a spatial membrane probability map (normalized to the integer range [$0$, $255$]). Synaptic clefts are treated largely the same, focusing on the darkened regions between a subset of adjacent cellular membranes. 

\begin{figure}[t]
	\begin{subfigure}[b]{0.22\textwidth}
		\centering
		\includegraphics[width=1.6in]{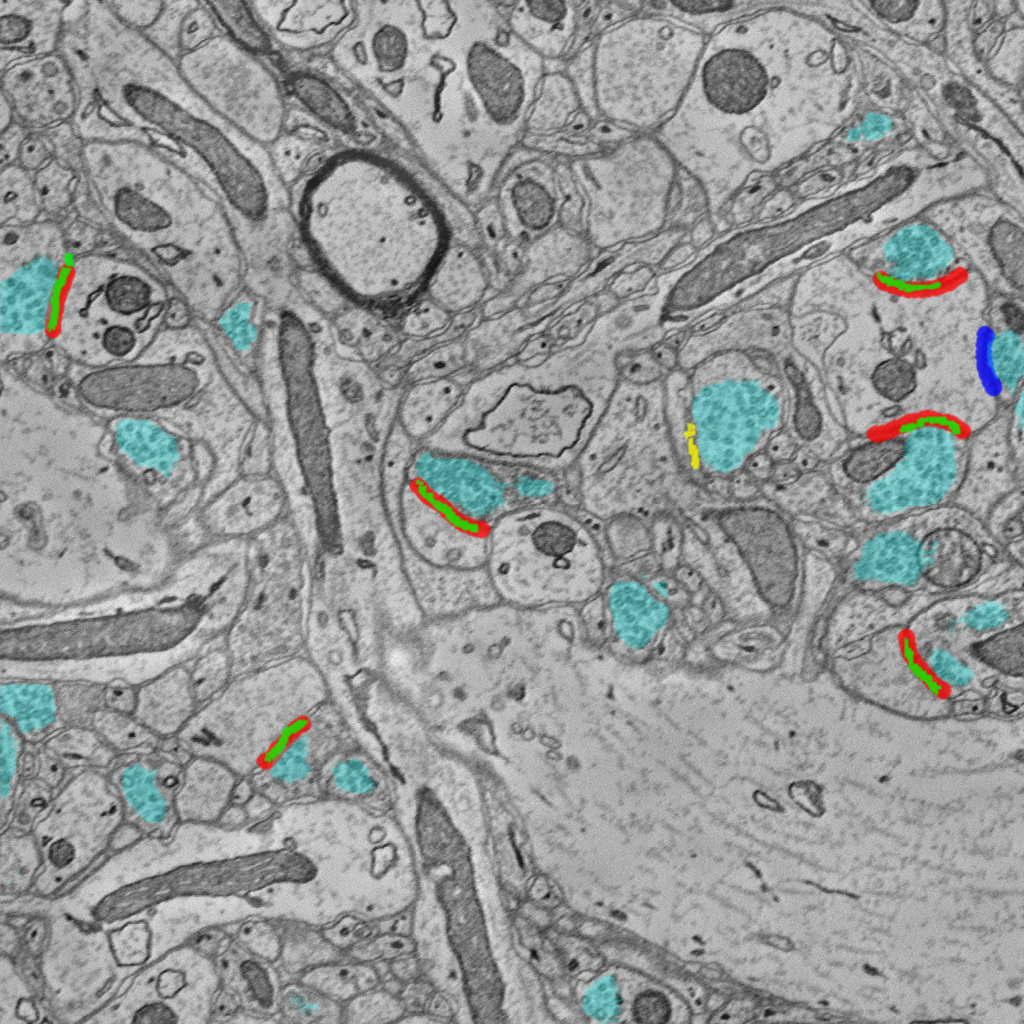}
		\caption{}
		
	\end{subfigure}
	\hfil
	\begin{subfigure}[b]{0.22\textwidth}
		\centering		
		\includegraphics[width=1.6in]{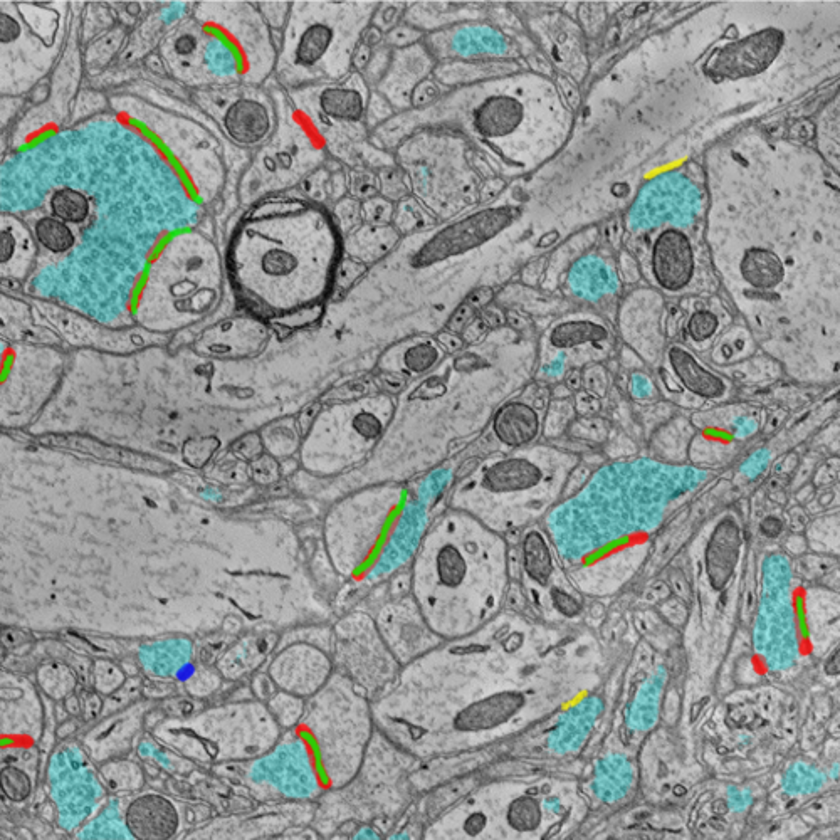}
		\caption{}
	\end{subfigure}
	\caption{Example of synapses and vesicles detected by our system, superimposed on raw electron microscopy samples. Examples show correctly identified ground truth synapses (red), missed ground truth synapses (blue), correct predictions (green), faulty predictions (yellow), and vesicle clusters (cyan).}
	\label{fig:block72}
\end{figure}

Our approach to vesicle detection is more unique. Vesicles are organelles that store the various chemicals required for neurotransmission, and are located at the pre-synaptic partner of neuron connections. Automatic detection of vesicles has previously been considered a difficult task~\cite{kreshuk2011automated} and has thus not been extensively studied. To our knowledge Vesicle-RF was the first system to attempt to identify individual vesicles (using a hand-crafted match filter) to assist with synapse identification~\cite{roncal2015vesicle}. Unlike their approach, we train a ConvNet to segment significant clusters of spatially co-located vesicles as a single feature (cyan in Figure~\ref{fig:block72}). In addition to allowing us to reconstruct directed connectivity by considering vesicle density in the pre- vs. post-synaptic partners, there is also a well-established correlation between this density and the strength of synaptic connections~\cite{morgan2016fuzzy} (akin to weights in an artificial neural net). 

\subsubsection{ConvNet Implementation}

\begin{figure}[t]
	\centering
	\includegraphics[width=3.25in]{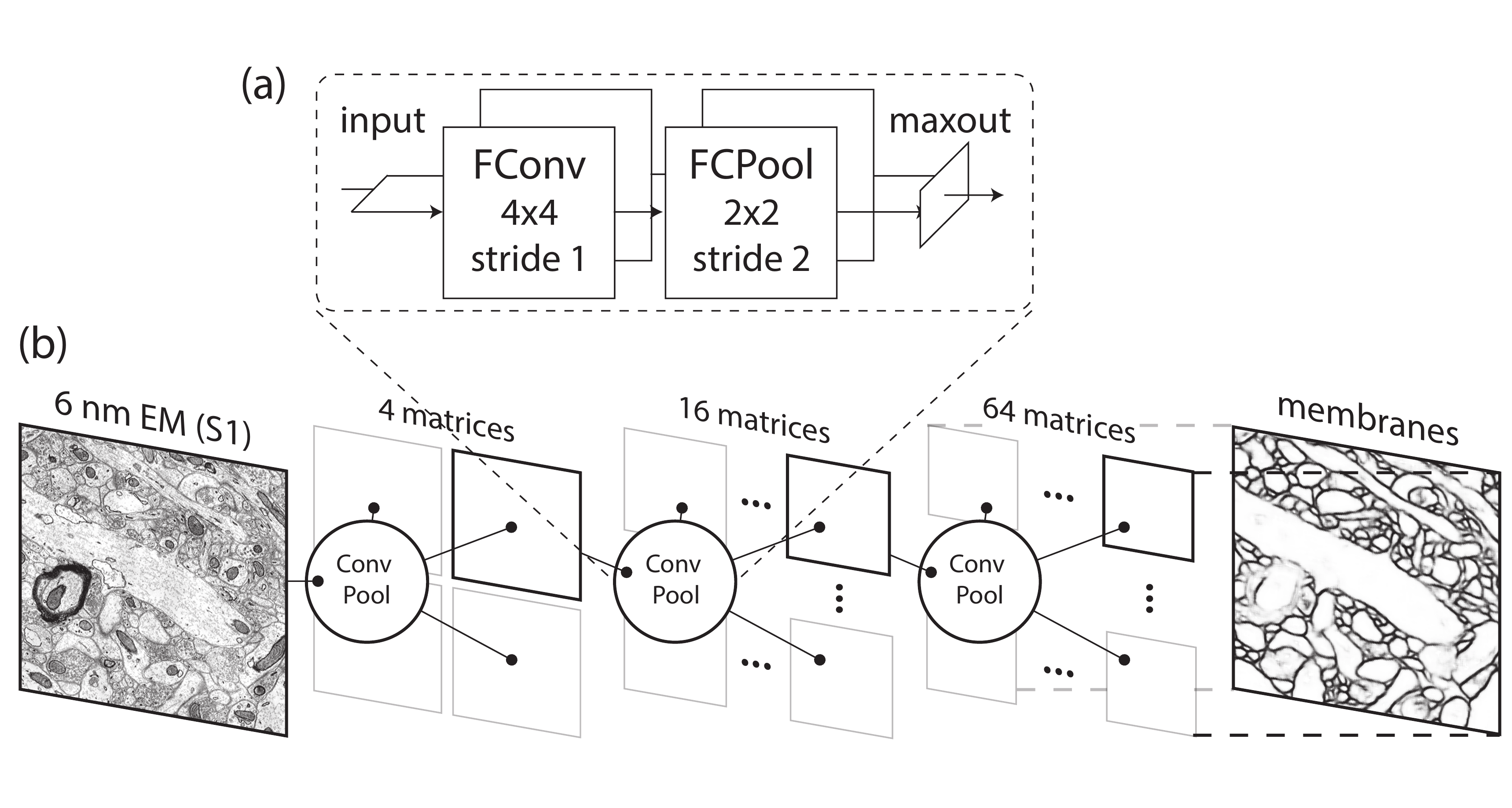}
	
	\caption{Lightweight ConvNet architecture used to detect synapses, vesicles and membranes. (a) Each ConvPool layer consists of alternating convolution/maxpool layers combined with a maxout function. (b) The full network comprises three consecutive ConvPool layers followed by a binary softmax. Our implementation is fully convolutional using the maxpooling fragments technique~\cite{giusti2013fast,masci2013fast}.}
	\label{fig:cnet}
\end{figure}

Each of our three marginal ConvNets adopts the same lightweight architecture (herein MaxoutNet, Figure~\ref{fig:cnet}) that comprises of consecutive ``ConvPool" modules. Each module consists of two parallel branches aggregated with a Maxout function \cite{seymour}, where each branch is a convolution operation (kernel size $k = 4\times 4$, 32 channels) followed by a $2\times 2$ maxpool of stride 2. The full network consists of three consecutive ConvPool modules followed by a final $6\times 6$ kernel to output the segmented feature map.

There are a number of compounding factors that allow our system to execute orders-of-substantially faster than previous state-of-the-art. First is the lightweight nature of the MaxoutNet architecture, which contains orders-of-magnitude fewer parameters than previous popular connectomics ConvNets (i.e., $\sim10^5$ compared to UNet's $\sim10^7$ \cite{ronneberger2015u}). Second is our minimally fully-convolutional implementation, inspired by fast algorithms of max-pooling fragments~\cite{giusti2013fast,masci2013fast}. Compared with naive patch-based methods of semantic segmentation (i.e., where the full convolutional field-of-view is reprocessed for every output pixel) this method leads to a $200-$fold reduction in redundant computations. Finally is our multicore CPU-optimized implementation, which makes liberal use of SIMD instructions~\cite{vanhoucke2011improving} and Cilk-based job scheduling~\cite{blumofe1996cilk} to yield $70-80$\% single-core utilization and $85$\% multi-core scalability across our $8-$core Haswell processor. The aggregate impact of these optimizations is an $11,000-$fold speed-up with respect to Vesicle-CNN \cite{roncal2015vesicle}. It is worth noting that Vesicle-CNN reports to be based on the N3 ConvNet architecture~\cite{ciresan2012deep} but uses stride-1 instead of stride-2 pooling, which is largely responsible for the inflation of their network size.

%Here, we would like to point out that while the architecture of Vesicle-CNN \cite{roncal2015vesicle} is supposed to be based on ``N3" \cite{ciresan2012deep}, they used 2x2 max-pooling with stride 1 (instead of 2 in the original ``N3"), so that the feature maps remain large. This choice leads to a huge expansion in the size of the first
%fully connected layer, which now maps $48$ $50 \times 50$ feature maps to $200$ $1\times 1$ feature maps, resulting in roughly 23M parameters. Both the accuracy and slow speed of this approach, are possibly a consequence of using a heavy ConvNet.

%\begin{figure}[t]
%	\centering
%	\includegraphics[width=3.25in]{Figures/david_graph}
%	\caption{(a) Comparison of performance of our approach framework with state-of-the-art approaches \cite{becker2013learning,roncal2015vesicle}  on the AC3 dataset \cite{roncal2015vesicle,wpd1}. Dashed lines indicate precision-recall curves obtained by varying post-processing parameters (see Section \ref{sec:comp}).  Solid lines correspond with the best F1-score for each. (b) Performance improvement obtained by considering only ``active" synapses (\emph{i.e.} those in proximity to identifiable vesicles). (c) Line graph F1-scores plotted as a function of synapse detection F1-scores. Our approach outperforms Vesicle-RF \cite{roncal2015automated} in graph accuracy, which is likely due to a combination of better segmentation and synapse detection accuracy.}
%	\label{fig:graph}
%\end{figure}

\subsection{Synapse Composition}
\label{sec:comp}

\begin{figure}[t]
	\centering
	\includegraphics[width=3.0in]{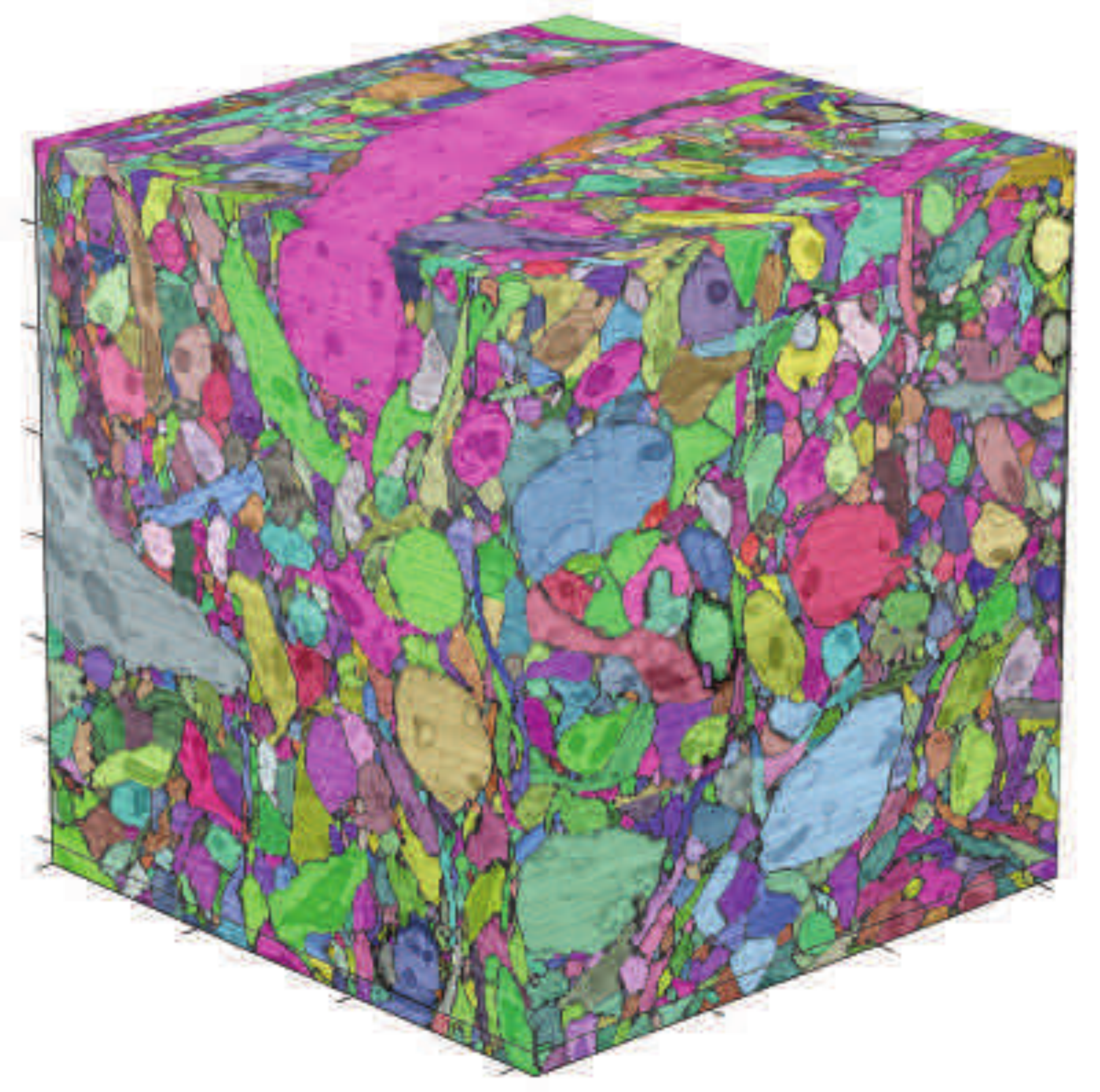}
	\caption{Example of neuron-level segmentation for a small sub-volume of the S1 dataset. This is reconstruction is achieved by (1) generating membrane segmentations with our lightweight marginal ConvNet, (2) producing an over-segmentation by applying the Watershed algorithm, and (3) merging adjacent segments with the Neuroproof package~\cite{parag2015context}. This segmentation is composed with intercellular clefts and synaptic vesicles to identify synapses.}
	\label{fig:cube}
\end{figure}

In recent years the ConvNet paradigm has been one of ``deeper is better", with more complex problems being solved by increasing network depth and data volume~\cite{krizhevsky2012imagenet,simonyan2014very,szegedy2015going}. This is also the Vesicle-CNN approach to synapse detection \cite{roncal2015vesicle}, which segments synapses directly from EM without explicitly capturing any prior biological knowledge. Although this approach certainly has merit (as reflected in state-of-the-art performance on ImageNet and other image processing benchmarks~\cite{russakovsky2015imagenet}) it does not extend well to problems where real-time performance is a hard constraint. Connectomics is one such problem \cite{lichtman2014big}. If a biologically meaningful volume of neural tissue already takes months-to-years to image with modern multi-beam electron microscopy, we certainly do not wish to exacerbate this problem with post-processing that is orders-of-magnitude slower. Further, producing ground-truth annotation for EM data is done manually and is highly time-consuming, taking even expert neuro-scientists hundreds of hours. As a result, the volume of training data available is extremely limited and there is need to design algorithms which are resilient to this. This motivates our choice of shallow networks, to avoid overfitting associated with deep networks in data-constrained problems.

In our system, we leverage the explicit compositionality of synapses to greatly reduce the search space within which our ConvNets are optimized (i.e. learning marginal versus joint distributions over features). This reduction in search space is explicitly captured by a lightweight network architecture, and thus able to be leveraged for substantial speed-up in synapse segmentation. To accomplish this, we take motivation from the concept of compositional hierarchies, wherein low-level features are explicitly composed to perform tasks such as object recognition~\cite{fidler2007towards,jin2006context,mjolsness1990bayesian,ullman2007object} and scene parsing \cite{farabet2013learning,zhao2011image}. It has been shown that in various recognition tasks, the use of hierarchies is more informative than non-hierarchical representations \cite{epshtein2005feature} and yields improved performance and generalizability, especially in the presence of occlusion or clutter \cite{lee2009convolutional,zhu2006hierarchical}. Such compositionality is also believed to be an intrinsic mechanism for object recognition in the visual cortex \cite{EJN530,riesenhuber1999hierarchical}. 

A key shortcoming of traditional compositional models is the difficulty of hand-crafting feature hierarchies for the expansive domain of natural images, or building suitable models to learn these features in an unsupervised fashion. This is one of the key motivators behind recent deep learning trends, although ConvNets are likewise criticized for their absence of compositionality \cite{fodor1988connectionism} and feature intelligibility \cite{tabernik2015adding}. Instead, our approach distils the advantages of both domains. Rather than capturing low-level features using Gabor filters or SIFT features, we take motivation from the work of Ullman ~\emph{et al.} in natural image classification~\cite{ullman2007object} and explicitly compose pictorial features. %Unlike this work, our pictorial features are not crops of the raw EM but instead taken from the ConvNet-segmented features maps of membranes, clefts and vesicles. 

Unlike previous work, our pictorial features are not crops of the raw EM but instead taken from the ConvNet-segmented features maps of membranes, clefts and vesicles. Here, learning feature (membranes, clefts and vesicles) segmentation using ConvNets is easier than learning synapses because of two primary reasons\textemdash (1) relative simplicity of features as compared to synapses (2) availability of significantly more ground truth for features (for \eg, the number of membrane pixels in an annotated EM stack is a few orders of magnitude larger than the number of synaptic pixels due to the sparsity of neural connections).  

Neuron membranes predicted by the ConvNet are further processed to segment individual neurons in the traditional manner~\cite{roncal2015automated}. Specifically, we generate an over-segmentation of neurons using the popular Watershed algorithm. These segments are then agglomerated using the NeuroProof package by Parag \emph{et al.}, which applies a pre-trained random forest classifier to determine which adjacent segments require merging~\cite{parag2015context}. In both cases we use custom implementations optimized for multicore CPU scalability~\cite{Victorshed,QuanNP}. An example output of this process is presented in Figure~\ref{fig:cube} for a small sub-volume of the S1 dataset. Neuron segmentations are then composed with cleft and vesicle segmentations to filter putative synapse candidates. This involves applying the following rules:
%, the execution time of which are negligible with respect to our optimized ConvNet implementation:

%%%%
\begin{enumerate}  \itemsep0em
	\item Restrict candidates to membrane regions and discard intra-cellular predictions 
	\item Constrain predictions to be within a certain maximum distance to vesicles. We are interested in finding active synapses typically marked by the presence of vesicles
	\item Threshold (binary) synapse probabilities with an empirically tuned parameter. % ($p_{T} =  0.7$).  
	\item Break large synapse predictions that cover many true and false synapses. This ensures each connected synapse candidate lies between a unique pair of neurons. This is important to find pre- and post-synaptic neurons for a synapse.
	\item Connected component analysis and checks on minimum 2D/3D size and slice persistence (similar to Vesicle-RF).
	\item Reject candidates that do not contain vesicles in one of the two neuron segments they connect. This additional check ensures that all the final synapse candidates are active and contain vesicles in close proximity, in one of the two adjoining segments.
\end{enumerate}

\begin{figure}[!t]
	\begin{subfigure}[t]{0.15\textwidth}
		\centering
		\includegraphics[height=1.85in]{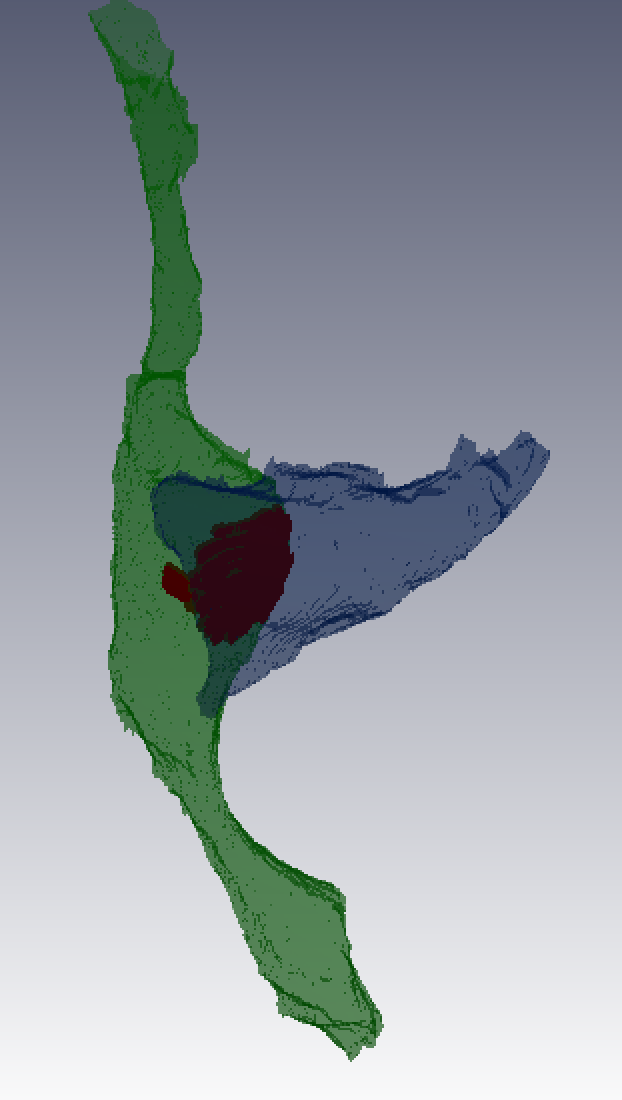}
		\caption{}
		
	\end{subfigure}
	\hfil
	\begin{subfigure}[t]{0.14\textwidth}
		\centering
		\includegraphics[height=1.85in]{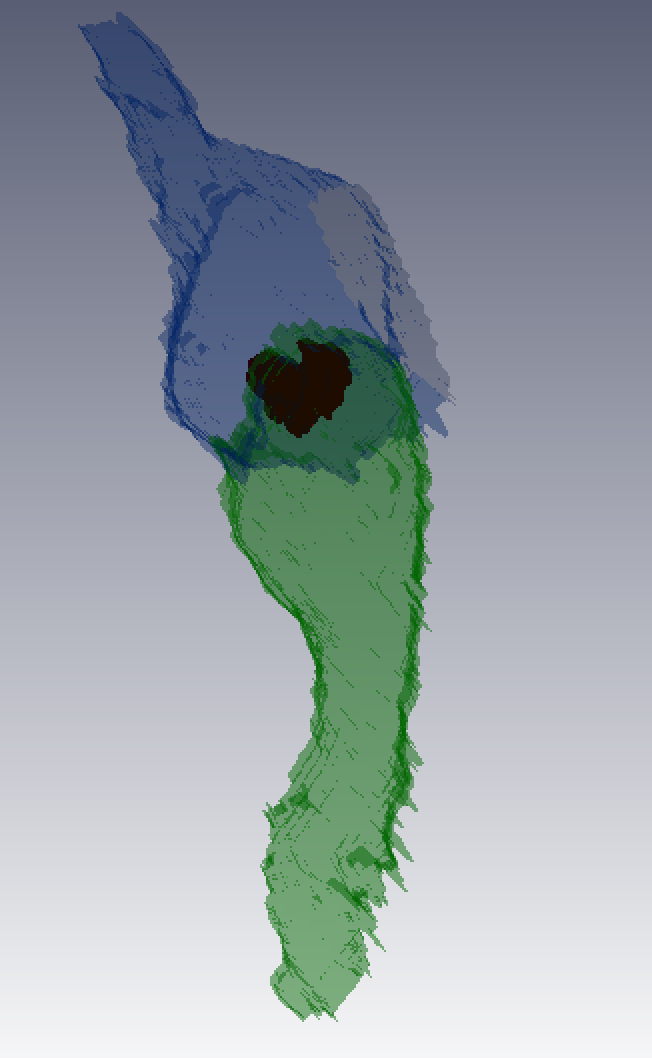}
		\caption{}
		
	\end{subfigure}
	\hfill
	\begin{subfigure}[t]{0.16\textwidth}
		\centering
		\includegraphics[height=1.85in]{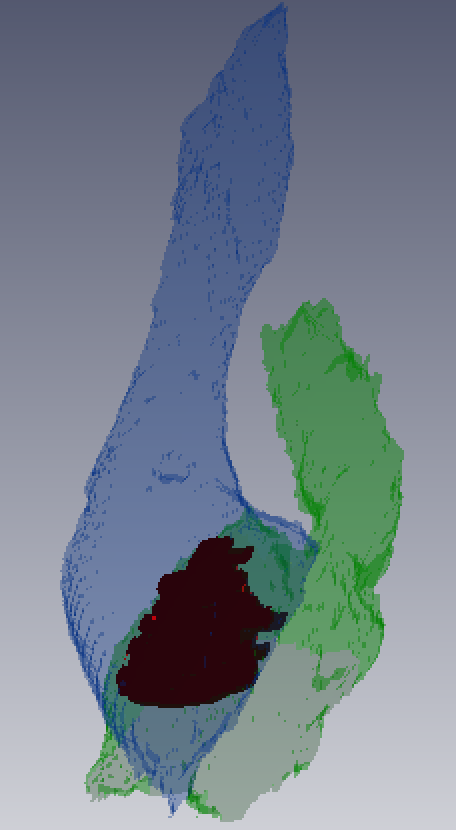}
		\caption{}
		
	\end{subfigure}
	%	\hfill
	%	\begin{subfigure}[t]{0.15\textwidth}
	%		\centering
	%		\includegraphics[height=1.5in]{Figures/nr80}
	%		\caption{}
	%		
	%	\end{subfigure}
	\caption{Automatic reconstruction of synapses from AC3 region of mouse S1. The axon/pre-synaptic neuron (green) and the dendrite/post-synaptic neuron (blue) interface via the synaptic cleft (red). The compositional nature of our system greatly simplifies the extraction of directionality.}
	\label{fig:neuronsAC3}
\end{figure}
%%%%
A few examples of synapses (red) detected using our system are presented in Figure \ref{fig:neuronsAC3}. The axon (pre-synaptic partner, green) and dendrite (post-synaptic partner, blue) are trivially identified from asymmetric vesicle density in our compositional model. Extracting directionality from Vesicle-RF/CNN or similar approaches would require additional layers of post-processing. As a result, our approach is the first to enable scalable inference of the the wiring diagram (defined as a graph connecting pairs of pre- and post-synaptic neurons with an inter-cellular synapse) without need for further processing or manual intervention.
\section{System Evaluation}
\label{sec : results}

We chose to evaluate our system performance, both in terms of reconstruction accuracy and execution time, on the largest available connectomics dataset. Specifically, the S1 dataset by Kasthuri \emph{et al.} is an anisotropic EM volume of mouse somatosensory cortex imaged at $3\times3\times29\,$nm resolution, containing an estimated $0.5-1$ synapses$/um^3$ \cite{busse2013automated}. This dataset is color-corrected and down-sampled to $6\times6\times29\,$nm prior to processing.

We compare our performance against the two state-of-the-art systems for synapse detection by Roncal \emph{et al.} -- the highly accurate ConvNet-based Vesicle-CNN and the faster but less accurate, random forest-based Vesicle-RF~\cite{roncal2015vesicle}. We select two independent $1024 \times 1024 \times 100$ volumes of S1\ for testing (AC3) and training/validation (AC4). Ground-truth annotations for these sub-volumes have been provided by expert neuroscientists~\cite{oc1}. 

\subsection{Evaluation Metrics}

Various metrics of pixel-level accuracy have been presented for various image segmentation challenges~\cite{csurka2013good}. However, in the connectomics scenario we are less interested in the pixel-level fidelity of synapse segmentation (i.e. the metric used for training and evaluating ConvNets) as long as we are correctly identifying the existence of a synaptic connection. This is convenient as it allows us to deploy far lighter networks without reducing the accuracy of the downstream connectivity graph.

For our benchmarking we consider the following metrics prescribed by IARPA for the MICrONs project~\cite{iarpa1} and used widely within the connectomics community~\cite{roncal2015vesicle,roncal2015automated}:

\begin{enumerate}
	\item\textbf{Synapse Detection Accuracy:} Precision, recall and F1-score (harmonic mean of precision of recall) of identified synapses with respect to neuroscientist-annotated ground-truth.
	
	\item\textbf{Wiring Diagram Accuracy:} Evaluating neuron connectivity graphs is complicated by the fact that our and earlier systems are simultaneously identifying both nodes (neuron segmentation) and edges (synapse identification). In order to evaluate the latter it is thus easier to consider the dual graph (herein ``line graph"), where synapses are captured as nodes and neuron segments as edges. The correspondence between nodes in the inferred line graph with respect to ground-truth can then be easily determined by assigning matching labels to spatially overlapping synapses. Further, both graphs are augmented to include all synapses present in the other to ensure an equal number of nodes.
\end{enumerate}
\begin{figure}[t]
	\centering
	\includegraphics[width=2.55in]{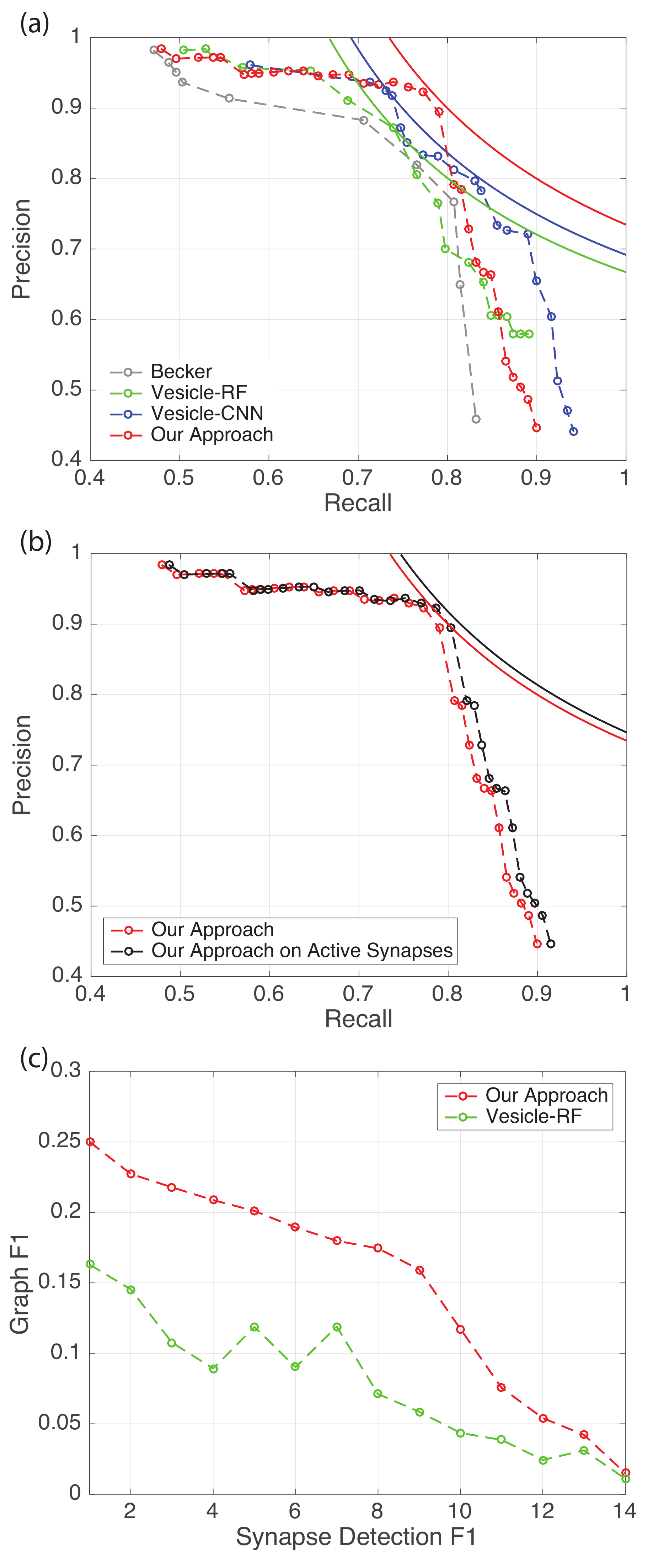}
	\caption{(a) Reconstruction accuracy of our approach system compared with state-of-the-art \cite{becker2013learning,roncal2015vesicle} on the AC3 S1 sub-volume dataset \cite{wpd1}. Dashed lines capture precision-recall curves and solid lines highlight the best operating point (based on F1-score) for each. (b) Performance improvement obtained by considering only ``active" synapses, i.e. those exhibiting a sufficient vesicle density. (c) Line graph F1-scores plotted against synapse detection F1-scores. Our approach outperforms Vesicle-RF \cite{roncal2015automated} in graph accuracy, indicating a combination of better segmentation and synapse detection accuracy.}
	\vskip -0.3in
	\label{fig:graph}
\end{figure}
\begin{table}[t]
	\centering
	\caption{Comparison of best operating points (F1-score) for our system compared to Vesicle-RF and Vesicle-CNN.}
	\label{tab: acc}
	\begin{tabular}{c|c|c|c|c}
		\textbf{Framework} & \textbf{Precision} & \textbf{Recall} & \textbf{F1}    & \textbf{Graph-F1} \\ \hline
		Our System         & \textbf{0.924}     & \textbf{0.782}  & \textbf{0.847} & \textbf{0.25}\\
		Vesicle-RF         & 0.89               & 0.71            & 0.790          & 0.16              \\
		Vesicle-CNN        & 0.917              & 0.739           & 0.817          & -                 
		
	\end{tabular}
\end{table}
\subsection{Synapse Identification Accuracy}
\label{ss:accuracy}
Figure~\ref{fig:graph}(a) presents the precision and recall of our synapse detection compared to previous state-of-the-art approaches~\cite{becker2013learning,roncal2015vesicle,wpd1}. The best operating point (based on F1-score) for each approach is summarized in Table~\ref{tab: acc}. Operating points with high recall are desirable for minimizing false negatives. Overall our approach attains a $7\%$ higher recall and $5\%$ higher F1-score than the previous best practical solution, Vesicle-RF. It also improves over the accuracy of Vesicle-CNN, which is prohibitively slow for non-trivial EM volumes.

In Figure~\ref{fig:graph}(b) we explore the impact of considering only active synapses (marked by a density of vesicles in the pre-synaptic partner). As discussed in Section~\ref{sec:comp}, it is believed that these synapses form the functional pathways of the connectome. Our results show that discarding inactive synapses improves the best operating point from an F1-score of $0.8470$ to $0.8627$.

Using the dual graph approach described above, we generate two line graphs $\mathscr{L}_{GT}$ (based on ground-truth) and $\mathscr{L}_{Pred}$ (predicted from our system). These graphs can then be compared to generate an additional set of precision, recall and F1-scores, which is an aggregate score capturing mistakes in both neuron segmentation and downstream synapse identification. A comparison of synapse and graph-level F1 scores is presented in Figure~\ref{fig:graph}(c) for our system and Vesicle-RF. Vesicle-CNN is not included as extracting a full connectome from the S1 volume would take an intractably long time.

We also explored if our rules-based composition module could be replaced with a ConvNet to improve detection accuracy. Specifically, we replaced the rules outlined in Section~\ref{sec:comp} with an extra ConvNet using the same MaxoutNet architecture as the marginal classifiers. We observed a peak F1-score of $0.467$ using this approach, which is a substantial step down from the $0.847$ reported above. We believe that this is a consequence of over-fitting to the very limited set of manually-annotated EM data. Increasing network depth did not improve performance.

\subsection{System Performance}

We compare the performance of our synapse detection system to the previous state-of-the-art Vesicle-RF, both theoretically (number of computations) and empirically (execution time). Benchmarking was conducted for the AC3 S1 sub-volume on an $8-$core Intel Core i7-5960X CPU with $32$ GB RAM. These results are summarized in Table~\ref{tab: perf}. We do not compare to Vesicle-CNN, which was shown by Roncal~\emph{et al.} to be $200-$fold slower than Vesicle-RF, and thus impractically slow for any non-trivial EM volumes. It is evident that the total number of instructions for Vesicle-RF and our approach is approximately the same ($\approx 6 \times 10^{12}$), although there is a noteworthy reduction in our execution time owing to our efficient multicore implementation.

\begin{table}[t]
	\centering
	\caption{Comparison of computation and speed of our approach to state-of-the-art Vesicle-RF. The time taken for membrane segmentation ($1.6$ hours versus $3$ weeks in prior work~\cite{roncal2015vesicle}) is discarded as sunk cost for fairer comparison.\\\hspace{\textwidth}}
	\label{tab: perf}
	\begin{tabular}{c|c|c|c|c}
		\textbf{Method}          & \textbf{Task} & \textbf{\#Instructions} & \textbf{Time (s)} & \textbf{CPU} \\ 
		\textbf{}          & & \textbf{  ($\times10^9$)} & & \textbf{ Use} \\ \hline
		{Vesicle-} & Vesicle                      & 379                              & 39.65             & 2.27                     \\
		{RF} & Synapse                      & 6,434                            & 1511              & 1.08                     \\ \hline
		{Our} & Vesicle                  & 2,863                            & 41.40             & 15.3                     \\
		{System} & Cleft                  & 2,811                            & 40.63             & 15.2                     \\                    
		& Rules                & 698                              & 78.82             & 1.42              \\                                                 
	\end{tabular}
\end{table}

Table~\ref{tab: perf} presents the following for each phase of the synapse detection: 
\begin{enumerate}
	\item \textbf{Instruction Count:} The number of instructions executed on the CPU 
	\item \textbf{Time:} The measured execution time
	\item \textbf{CPU count:} The average CPU utilization with respect to theoretical maximum throughput per thread (maximum of 16 threads)
\end{enumerate}

The performance of a connectomics system can only be truly assessed when deployed on the large-scale datasets made possible by recent advances in microscopy. Moving beyond the AC3 sub-volume, we use our system to perform a reconstruction of the full 245 GB S1 dataset and compare to a time for Vesicle-RN/CNN based on the authors' reported findings~~\cite{roncal2015vesicle,roncal2015automated}. Our system took a total of 9.7 hours to reconstruct the full dataset, from EM to connectivity graph. The majority of this time was spent by the ConvNets responsible for segmenting membranes, clefts and vesicles, with each network taking $\sim 1.8$ hours followed by $\sim 1.7$ hours of post-processing. If we ignore the time required for membrane segmentation (as per the Vesicle-RF study, where this step took 3 weeks on a farm of 27 Titan X GPUs~\cite{roncal2015automated}) then our total processing time reduces to 8.1 hours for 245 GB. This is approximately $20$ times faster than Vesicle-RF ($39\,$hours$/56\,$GB) and $11,000$ times faster than Vesicle-CNN ($370\,$hours$/$GB, extrapolated from AC3 reconstruction time), both running on a $100$ core CPU cluster with $1\,$TB RAM.

Restating these results, even if one ignores the time required for membrane segmentation, detecting synapses in the full S1 volume would take approximately 1 week with Vesicle-RF or 12 years with Vesicle-CNN. As demonstrated in Section~\ref{ss:accuracy}, our system is more accurate and orders-of-magnitude faster than either approach. 

\section{S1 Reconstruction}

\begin{figure}[t]
	\centering
	\includegraphics[width=3.25in]{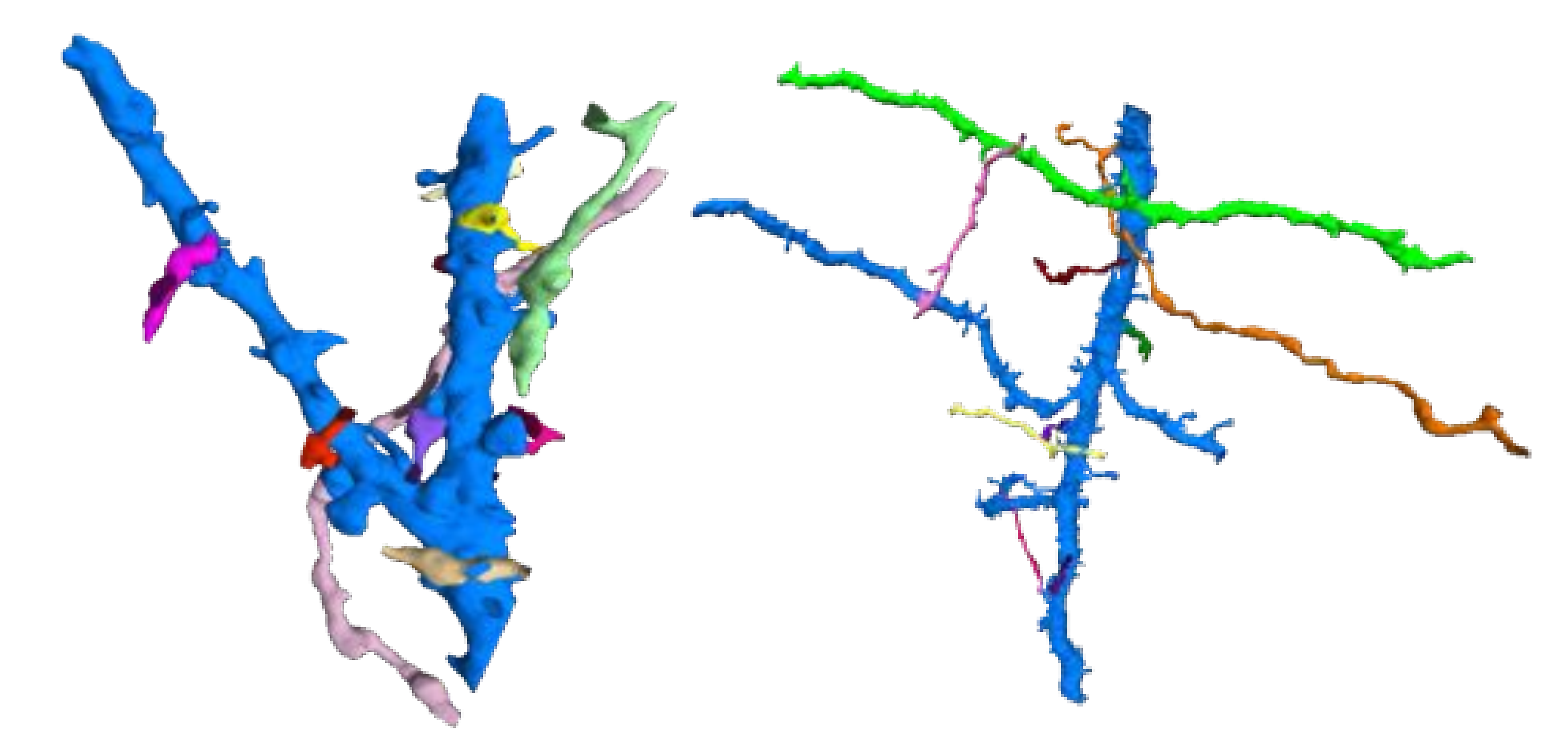}
	\caption{Rendering of automatic reconstructions from S1 obtained with our approach: Shown is a dendrite (blue) and its corresponding innervating axons (uniquely colored).}
	\label{fig:neurites}
\end{figure}

To our knowledge, the only previous attempt to reconstruct a non-trivial sub-volume of S1 was by Roncal \emph{et al.} using Vesicle-RF. The authors reported that ${11,489}$ synaptic connections were detected in a ${60,000\, \mu m^3}$ volume of S1, obtained by inscribing a $6000 \times 5000 \times 1850$ cube ($56\,$GB of data) within the down-sampled $6\times 6\times 29\,$nm version of S1~\cite{roncal2015automated}. It is unclear how this result relates to the $50,335$ reported for the same volume and method in \cite{roncal2015vesicle}. Applying our system, we identify ${66, 162}$ synapses in ${95, 102\, \mu m^3}$, corresponding to non-black pixels in $\approx\,110\,$GB of non-corrupt EM data extracted from $\approx\,245\,$GB of raw EM. It is difficult to compare these results directly since the relationship between Roncal's and our data remains unclear -- \eg S1 contains large regions of soma and blood vessels that do not contain any synapses. These results are equivalent to a synapse density estimate at $0.695$ synapses$/\mu m^3$, which corresponds to $0.889$ synapses$/\mu m^3$ if we factor in the 78.1\% recall of our method. These numbers agree with projections of S1 synapse density ranging from $0.5-1$ synapses$/\mu m^3$ \cite{busse2013automated}. 

The remainder of this Section demonstrates different spatial scales at which our automated reconstructions can be investigated. Figure \ref{fig:neurites} shows a reconstruction of select dendrites, axons and their synapses from S1.  While some axons synapsing to the dendrite (blue) span the entire volume, several of them are fragmented. This is due to split errors in the segmentation and emphasizes that synaptic-level reconstruction of EM data goes hand-in-hand with neuron segmentation. The two should be used to iteratively refine one another in future attempts to accurately reconstruct larger-scale connectomes. To the best of our knowledge, such automated reconstructions of dendrites and their innervating axons have never been seen before; generating similar visuals would require time-intensive manual annotation.

\begin{figure}[t]
	\centering
	\includegraphics[width=3.0in]{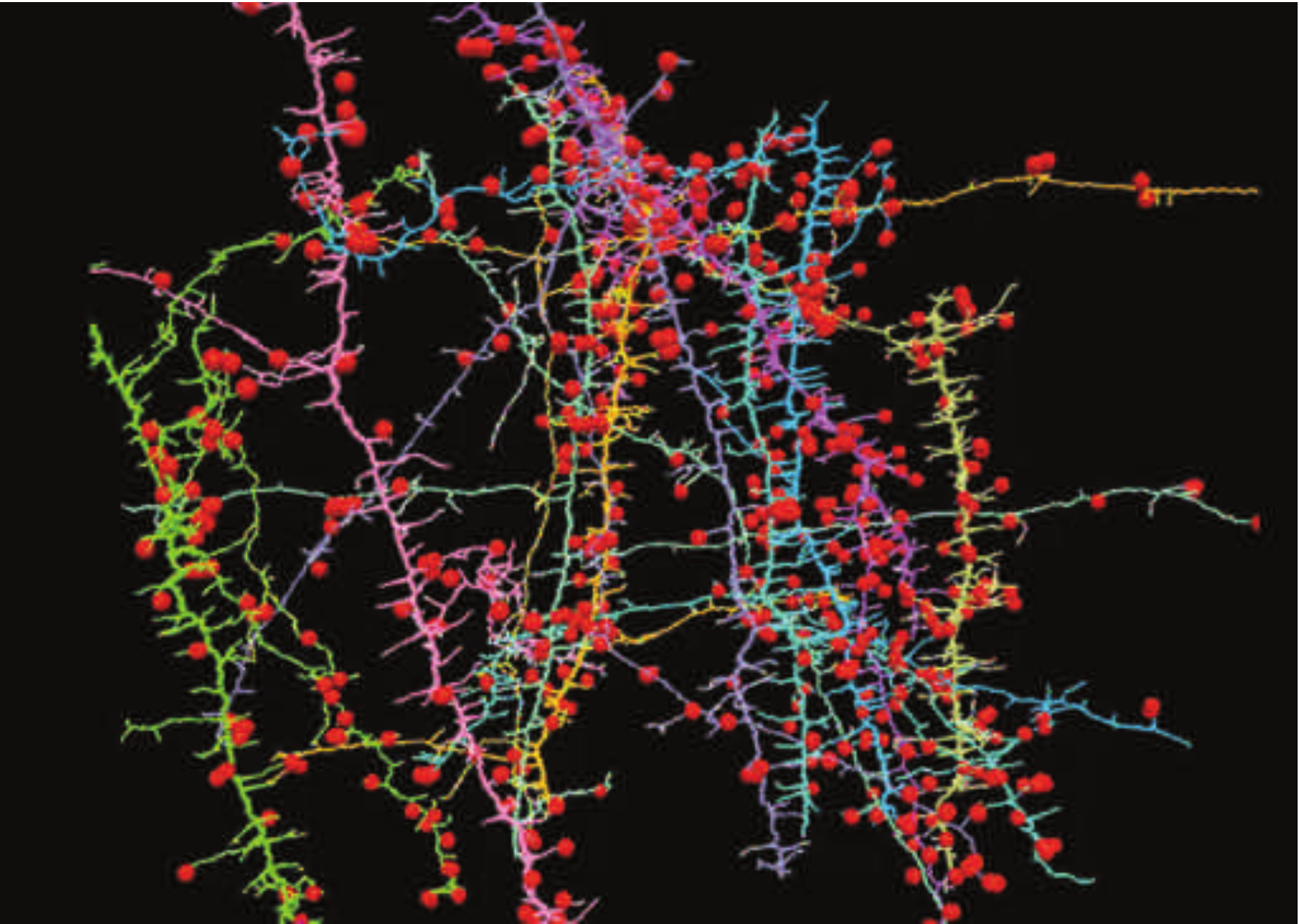}
	\caption{Nine neurite skeletons sampled from our automated S1 reconstruction (uniquely colored). The 653 associated synapses detected by our system are overlaid in red. }
	\label{fig:synapses}
\end{figure}

\begin{figure}[t]
	\centering
	\includegraphics[width=2.9in]{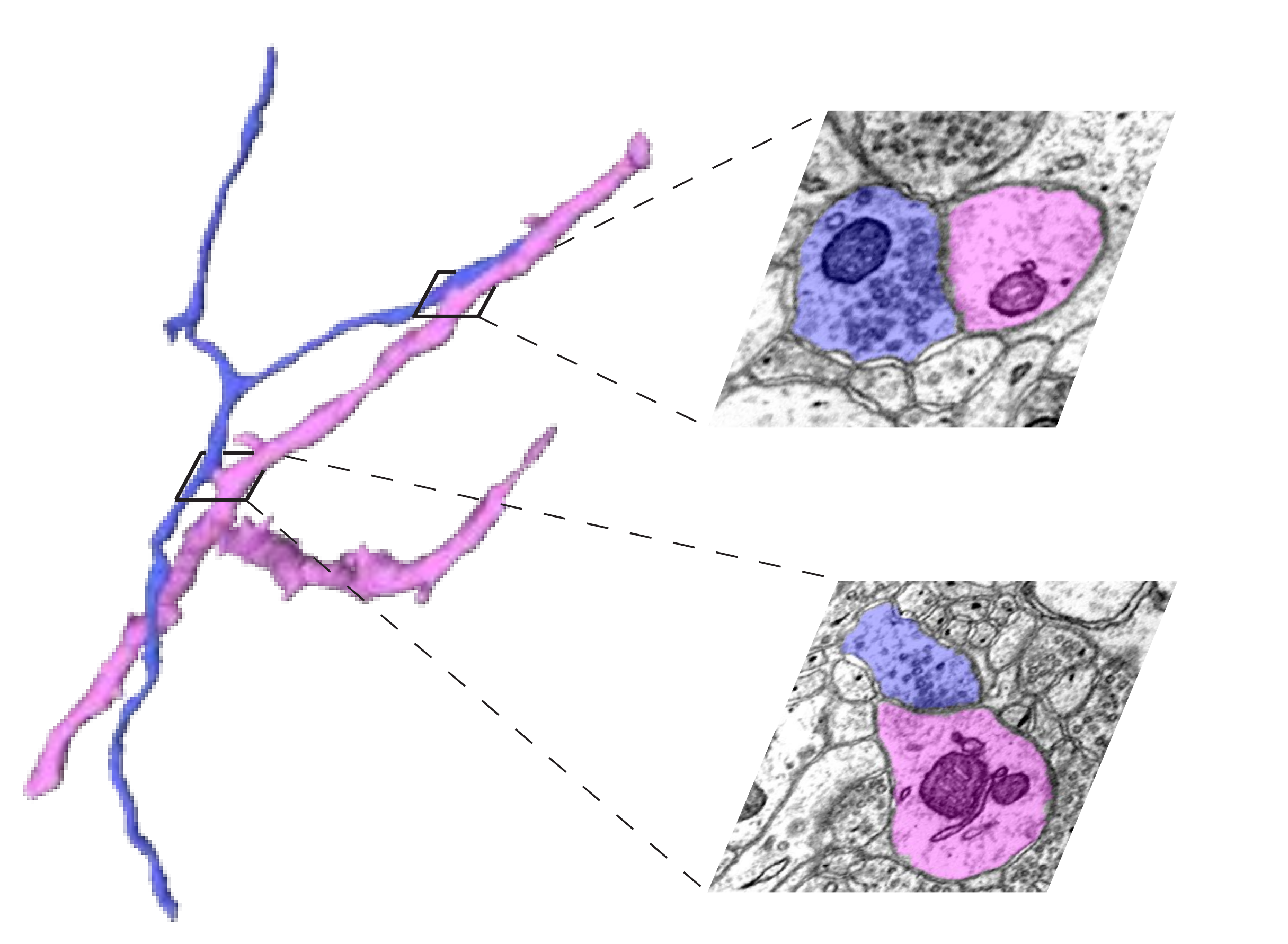}
	\caption{Automatic detection of an axon (blue) from S1 innervating a dendrite (pink) at two different locations. Multiple ``redundant" inhibitory connections are yet to be studied, highlighting the biological insights that can be gained from the connectomics field.}
	\label{fig:2 synapse}
\end{figure}

In Figure~\ref{fig:synapses} we zoom out to inspect 9 neurite skeletons extracted from the S1 reconstruction. The 653 associated synapses detected by our system are overlaid in red, showing the overall density and spatial distribution. Even without further data analysis this style of visualization can provide novel biological insight. Consider Figure \ref{fig:2 synapse}, where we see an axon (blue) synapsing the same dendrite at two separate locations. Since the axon is innervating the dendrite along the shaft and not a spine, it is likely to be an inhibitory axon. Multiple redundant inhibitory connections are yet to be studied and have previously not been detected using automated methods (discussions with Jeff Lichtman \cite{kasthuri2009brain}). This result highlights the power of automatic reconstruction in rapidly revealing insights into brain morphology.

\section{Discussion}
\label{sec: discussion}

Reconstructing large-scale maps of neural connectivity is a critical step toward understanding the structure and function of the brain. This is an overarching goal of the connectomics field, and although conceptually straightforward, there are unprecedented ``big data" issues that need to be overcome for investigating non-trivial tissue volumes~\cite{lichtman2014big}. With state-of-the-art multi-beam electron microscopes, a 1 $mm^3$ volume (comprising several petabytes of data) can be imaged within months and is thus an attainable goal. It is critical that the computational aspect of a connectomics pipeline operate at a similar pace in order to facilitate large-scale investigation of neuron morphology and connectivity.

Mirroring recent trends in the wider computer vision community, deep ConvNets have been adopted as a de facto standard for the complex image segmentation tasks required for connectomics~\cite{ronneberger2015u, roncal2015vesicle, lichtman2014big}. Good progress has been toward the segmentation and morphological reconstruction of neurons, with a recent study presenting a multicore CPU system capable of operating within the same order-of-magnitude as microscope-pace~\cite{matveev2016}. However, there has been relatively little success in large-scale identification of the synaptic connections between neurons, i.e., the ``edges" in a connectivity graph. These features are more difficult to identify owing to both (a) their small size and complex compositionality, and (b) their comparative underrepresentation in manually-annotated data. The best existing ConvNet solution (Vesicle-CNN) would take many years to reconstruct the 245 GB S1 dataset, and even a less accurate random forest classifier (Vesicle-RF) takes more than a week~\cite{roncal2015vesicle}.

In this study we have presented the first practical solution to high-throughput synapse detection, closing the gap toward microscope-pace by two orders-of-magnitude. Taking inspiration from previous research in compositional hierarchies~\cite{ullman2007object}, our system is comprised of two stages: (1) a bank of lightweight CNNs for segmenting ``marginal" features (i.e. membranes, intercellular clefts and synaptic vesicles), and (2) a rules-based model that explicitly composes these features based on prior biological knowledge. This system yields an improvement of 5-to-7\% compared to previous state-of-the-art, and we are interested to see whether this compositional ConvNet methodology can be applied to a broader class of image segmentation tasks.

Similar to Matveev et al.'s pipeline~\cite{matveev2016} for reconstructing neuron morphology (i.e. ``nodes" in a connectivity graph), we choose to optimize our implementation specifically for multicore CPU \cite{budden2016deep} systems to remove the bottleneck of data communication rates. Our implementation is $20-$fold faster than Vesicle-RF and $11,000-$fold faster than Vesicle-CNN, and we believe that this marks an important contribution toward the goal of a streaming connectomics pipeline for investigating the connectivity of biologically meaningful volumes of neural tissue.

\bibliographystyle{unsrt}
{\small
	\bibliographystyle{ieee}
	\bibliography{egbib}
}

\end{document}